\documentclass[journal]{IEEEtran}
\usepackage{xcolor}

\usepackage{cite}

\ifCLASSINFOpdf
  \usepackage[pdftex]{graphicx}
\else
  \usepackage[dvips]{graphicx}
\fi
\usepackage{amsmath}
\usepackage{amssymb}
\usepackage{array}

\usepackage{subcaption}
\usepackage{bm}

\newcommand{\vect}[1]{\bm{#1}} 
\newcommand{\icol}[1]{
  \begin{bmatrix}#1\end{bmatrix}^\intercal%
}

\newcommand{\msq}{\text{m}^2}
\newcommand{\mcb}{\text{m}^3}
\newcommand{\Nm}{\text{N·m}} 

\newcommand{\nR}[1]{\mathbb{R}^{#1}}		

\begin{document}
%
\title{Forbal: Force Balanced 2-5 Degree of Freedom Robot Manipulator Built from a Five Bar Linkage}
%
%
%

\author{Yash Vyas*
        and~Matteo~Bottin*
\thanks{*Department of Industrial Engineering, University of Padua, via Venezia 1, Padova (PD) 35131 - Italy. {\tt \footnotesize {Email: yashjanardhan.vyas@phd.unipd.it}. }}
\thanks{Manuscript version: \today.}}

%
%

\markboth{}%
{Shell \MakeLowercase{\textit{et al.}}: Bare Demo of IEEEtran.cls for IEEE Journals}
%



\maketitle

\begin{abstract}
A force balanced manipulator design based on the closed chain planar five bar linkage is developed and experimentally validated. We present 2 variants as a modular design: Forbal-2, a planar 2-DOF manipulator, and its extension to 5-DOF spatial motion called Forbal-5. The design considerations in terms of geometric, kinematic, and dynamic design that fulfill the force balance conditions while maximizing workspace are discussed. Then, the inverse kinematics of both variants are derived from geometric principles. 

We validate the improvements from force balancing the manipulator through comparative experiments with counter mass balanced and unbalanced configurations. The results show how the balanced configuration yields a reduction in the average reaction moments of up to 66\%, a reduction of average joint torques of up to 79\%, as well as a noticeable reduction in position error for Forbal-2. For Forbal-5, which has a higher end effector payload mass, the joint torques are reduced up to 84\% for the balanced configuration. Experimental results validate that the balanced manipulator design is suitable for applications where the reduction of joint torques and reaction forces/moments helps achieve millimeter level precision.
\end{abstract}


%
\IEEEpeerreviewmaketitle

\section{Introduction}
Robot manipulators are now increasingly part of industrial automation, forming the bedrock of modern manufacturing processes. These robots are typically used to perform repetitive and strenuous tasks, and are particularly important where consistent precision is required over long time periods. 
In recent years, there has also been research in mobile robot manipulation for applications such as construction, search and rescue, and household tasks \cite{GhodsianEtAl2023MobileManipulatorsIndustry}. Manipulators have been deployed in space for decades to carry out specialized measuring and maintenance tasks \cite{PapadopoulosEtAl2021RoboticManipulationCapture}.

Both serial (open-chain) and parallel (closed-chain) manipulators are employed in industrial applications. Whereas serial robots have a higher repeatability, parallel manipulators have higher stiffness and allow for higher accelerations and payloads than serial manipulators \cite{Merlet2006ParallelRobots}.

A drawback to both serial and parallel manipulators is the constant motion of the center of mass, which often extends far from the robot base attachment where the reaction forces are loaded. This offset of the gravity force, as well as joint torques and link accelerations, results in reaction forces and moments at the base, and requires joint torques for gravity compensation. The reaction forces and moments acting on the base can cause instability, and vibrations for fast motions. In addition, often this reduces precision, and increases energy consumption as well as component fatigue due to a large component of the joint torque being expended on gravity compensation.

This problem can be addressed through intelligent mechanical design that balances the manipulator for all motions. Various types of balancing exist, which can be clustered into two broad categories: balancing the gravity and other statically applied forces and moments \cite{Gosselin2008GravityCompensationStatic}, and balancing of dynamic forces and moments caused by motion \cite{ArakelianSmith2005ShakingForceShaking}:
\begin{itemize}
    \item \textbf{Gravity Balancing:} the static gravity force is balanced  along the gravity axis.
    \item \textbf{Static Balancing:} all static forces (e.g. gravity, and potential forces from springs) are balanced in all axes, so that the actuators do not need to apply torque at the equilibrium configuration.
    \item \textbf{Force Balancing:} dynamic forces from the motion of manipulator links are balanced so that the center of mass is static and the net force at that position is zero.
    \item \textbf{Moment Balancing:} dynamic moments about the manipulator's center of mass are zero.
    \item \textbf{Dynamic Balancing:} both force and moment balanced about the center of mass, so the reaction forces and moments are zero for all motions (also known as reactionless).
\end{itemize}

The field of balanced robotic manipulator design has seen significant progress in the past few decades, spurred by advances in theoretical methods to derive dynamic constraints for various kinematic arrangements \cite{LowenBerkof1968Surveyinvestigationsbalancing,Wijk2014Methodologyanalysissynthesis}, as well as experimental prototypes to demonstrate their performance. 

Static balancing can be achieved with the addition of low mass components such as springs \cite{Gosselin2008GravityCompensationStatic,AgrawalFattah2004Gravitybalancingspatial}. 
As springs exert forces when moved from the equilibrium configuration, dynamic forces and moments can only be balanced through kinematic and dynamic modifications. 
Parallel manipulators, such as the Delta Robot \cite{ClarkEtAl2022BalancedDeltaRobot}, are the preferred option for dynamic force and moment balancing due to lower mass increase \cite{WijkEtAl2011GenericMethodDeriving}. Here, the kinematic constraints of the mechanism are incorporated with the dynamic equations to derive balance conditions between geometric and dynamic parameters, known as inherent balancing. 

Inherent Balancing for dynamic force balancing requires modifications of the link masses and center of mass (CoM) positions, so that the CoM of the manipulator is static for all admissible motions. Applying this approach to moment balancing is more difficult, requiring precision manufacturing for both mass, CoM and inertia of each link so that the manipulator inertia is zero for admissible motions. Numerous novel designs have been developed for dynamic force balancing  \cite{BriotEtAl2009PAMINSAnewfamily, SuryavanshiEtAl2023ADAPT3Degrees} and dynamic moment balancing \cite{WijkEtAl2013Designexperimentalevaluation,ZomerdijkWijk2022StructuralDesignExperiments}. 

For industrial tasks, force balanced manipulators are suitable as they remove gravity compensation and dynamic force contributions to joint torques, which are the majority at low to medium operating speeds. Although incorporating moment balancing results in fully reactionless motion, it is more difficult, requiring precise link inertias, and is not easily adjustable to changing end effector payloads.

\begin{figure}[!ht]
    \centering
    \includegraphics[width=0.8\linewidth]{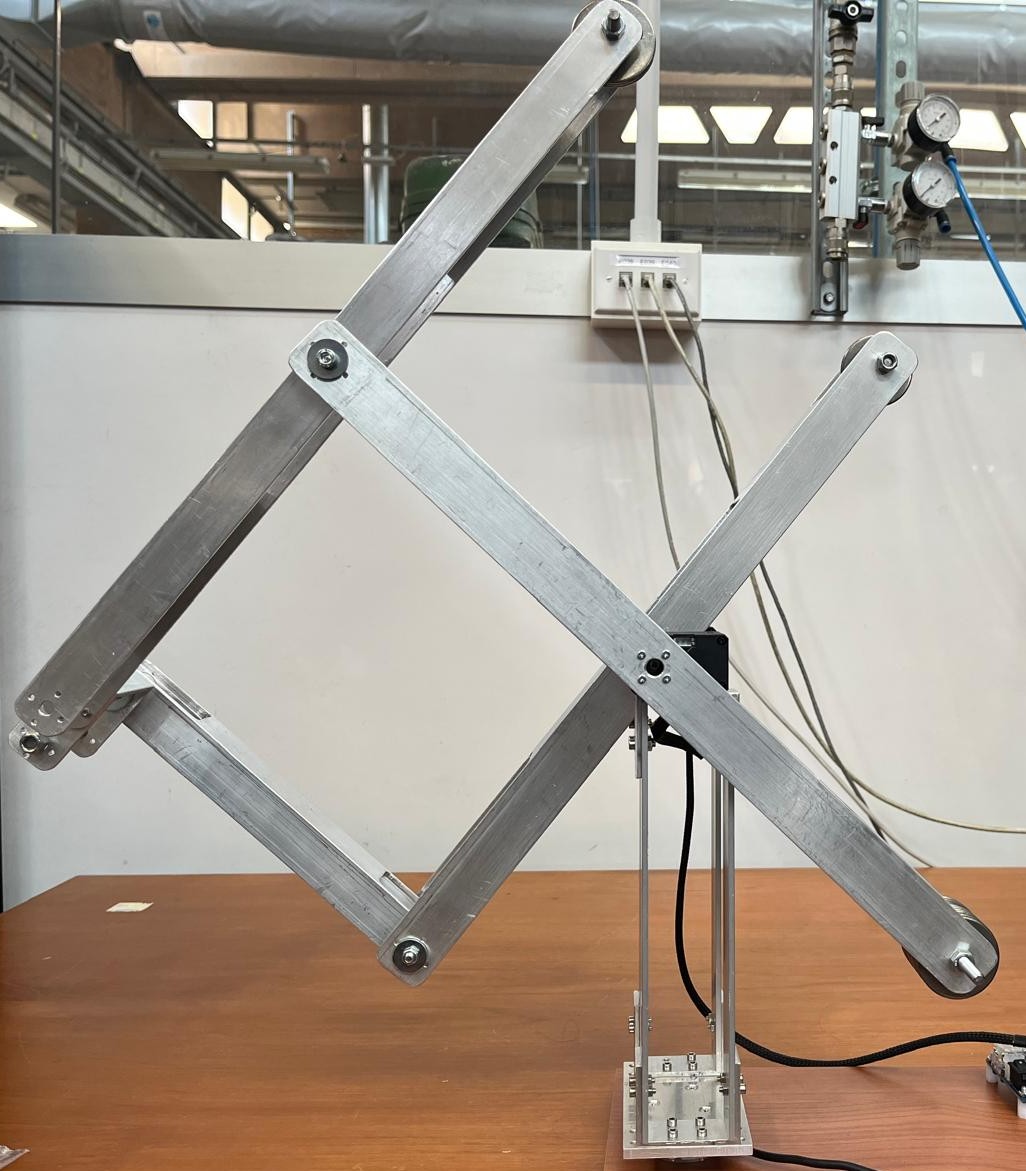}
    \caption{Forbal-5}
    \label{fig:forbal-2}
\end{figure}

\begin{figure}[!ht]
    \centering
    \includegraphics[width=0.8\linewidth]{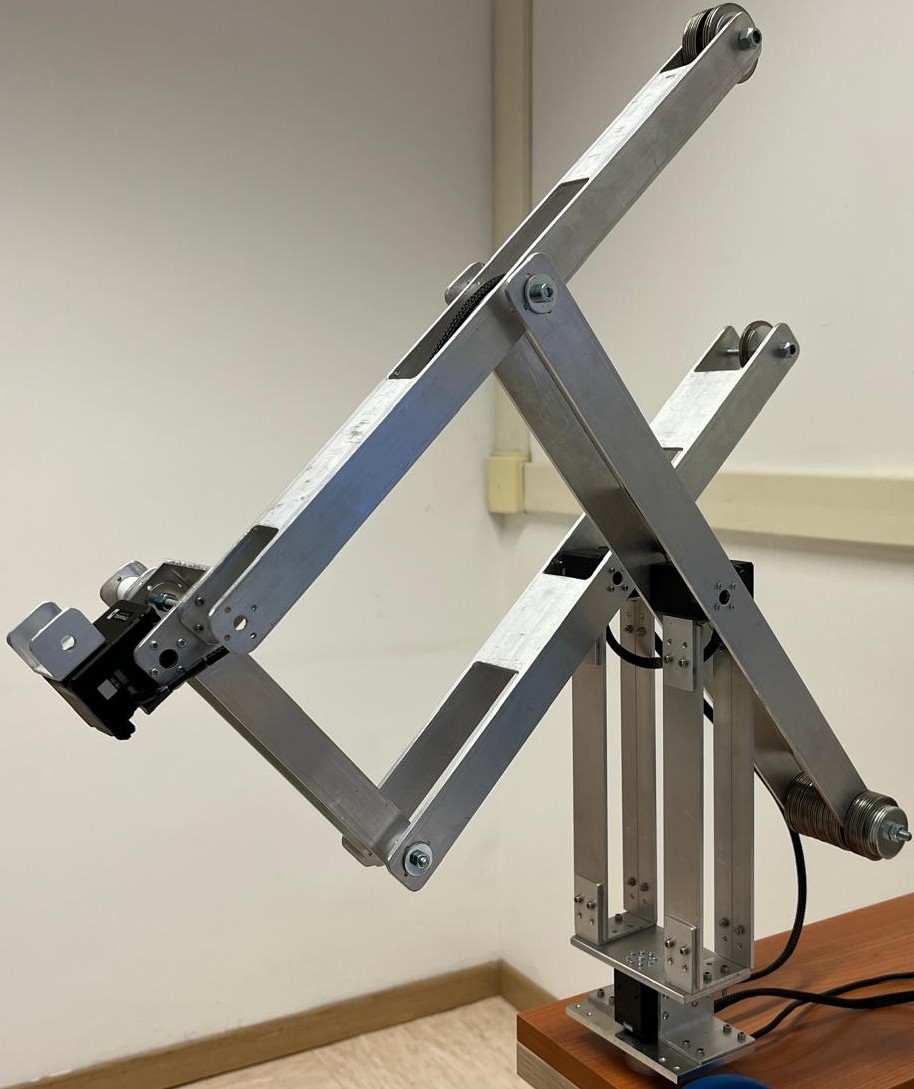}
    \caption{Forbal-5}
    \label{fig:forbal-5}
\end{figure}

For this reason, in this research we focus on force balancing for robot manipulation. This paper presents a novel \textbf{for}ce \textbf{bal}anced manipulator built out of a closed chain five bar linkage (5BL) design, named \textbf{Forbal}. We present two modular variants: 
\begin{itemize}
    \item \textbf{Forbal-2:} a 2 degree of freedom (DOF) planar manipulator that demonstrates the scientific principle and design innovation of force balancing (shown in Fig. \ref{fig:forbal-2})
    \item \textbf{Forbal-5:} extension of Forbal-2 with an additional 3-DOF, allowing for a full 3D position and 2 angle pose for the end effector while retaining force balance properties (shown in Fig. \ref{fig:forbal-5}).
\end{itemize}

We also derive the direct inverse kinematic controller for both variants using only the geometry and kinematics, which is a more computationally efficient method compared to nonlinear approaches. The built prototype of this manipulator in both variants is tested for various motion trajectories, comparing the improvements from force balancing on the reaction forces/moments, joint torques, and precision.

The contributions of this paper are as follows:
\begin{itemize}
    \item Mechanical design and control of a force balanced five bar linkage robotic manipulator with 2-5 degrees of freedom (DOF).
    \item Experimental validation of the reduction in reaction forces / moments, joint torques and improvement in precision from force balancing of the manipulator.
\end{itemize}

The organization of this paper is as follows. Section \ref{sec:mechanical_design} presents the force balance conditions for the five bar linkage, evaluates key design decisions, and presents the mechanical design for both variants. The inverse kinematic control for 2D position control of Forbal-2 and 5-DOF pose control of Forbal-5 is presented in section \ref{sec:ik_control}. 
Experimental validation of the force balancing properties and analysis of the impact of counter mass balancing are presented in section \ref{sec:experiments}. The research and its results are summarized in the Conclusion (section \ref{sec:conclusion}).
\section{Mechanical Design}\label{sec:mechanical_design}
\subsection{Force balance of a planar five bar linkage}\label{ssec:force_balance_derivation}
\begin{figure}[!ht]
    \centering
    \includegraphics[width=\linewidth]{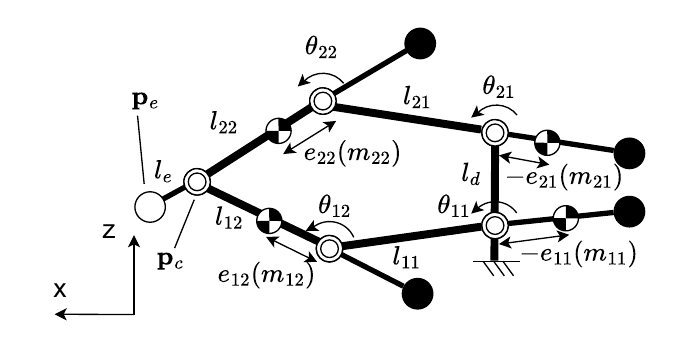}
    \caption{The general model of a force balanced five bar linkage manipulator.  The link length is \(l_{ij}\) and its center of mass vector along the length axis is \(e_{ij}\) with mass \((m_{ij})\). Counter masses (black circles) are mounted to adjust the link mass and CoM position. The end-effector is attached to Link 22. The closed loop position is \(\vect{p}_c\) and end-effector position is \(\vect{p}_e\).}
    \label{fig:fivebl}
\end{figure}

The force balance conditions for parallel manipulators composed of 2-4 RRR chains were derived by \cite{Wijk2014Methodologyanalysissynthesis}.
Here, we present a similar approach based on the Linear Momentum to derive the conditions for a planar five bar linkage placed on the vertical plane, as shown in Fig. \ref{fig:fivebl}. 

The five bar linkage is modeled as made of two RR chains that connect at $\vect{p}_c$. Joints and links are identified by a subscript $ij$, where $i$ denotes the chain number, and $j$ denotes the joint and connected link in that chain, starting from the base.
The first joint in each of the two chains is aligned with the $z$-axis of the fixed frame and placed at a height of $l_h$ on the base.

The linear momentum \(\vect{L}\) of the manipulator is:

\begin{equation}
    \vect{L} = m_{11} \dot{\vect{r}}_{11} + m_{12} \dot{\vect{r}}_{12} + m_{21} \dot{\vect{r}}_{21} + m_{22} \dot{\vect{r}}_{22} .
    \label{eq:linear_momentum_cart}
\end{equation}
where $m_{ij}$ and \(\dot{\vect{r}}_{ij}\) are the masses and the Cartesian linear velocity vector of the CoM of link \(ij\) about the fixed frame, respectively.

CoM velocities \(\dot{\vect{r}}_{ij}\) are derived from the forward kinematics and substituted into \eqref{eq:linear_momentum_cart}:

\begin{equation}
\begin{split}
    \vect{L} = (m_{11} e_{11} + m_{12} l_{11})
    \begin{bmatrix}
    -\sin(\theta_{11}) \\ \cos(\theta_{11})
    \end{bmatrix} &
    \dot{\theta}_{11} \\
    + (m_{21} e_{21} + m_{22} l_{21})
    \begin{bmatrix}
        - \sin(\theta_{21}) \\
        \cos(\theta_{21})
      \end{bmatrix} 
    & \dot{\theta}_{21} \\ + m_{12} e_{12}
    \begin{bmatrix}
        -\sin(\theta_{12}) \\
        \cos(\theta_{12})
    \end{bmatrix}
    \dot{\theta}_{12} + m_{22} e_{22}
    \begin{bmatrix}
        - \sin(\theta_{22}) \\
        \cos(\theta_{22})
    \end{bmatrix} & \dot{\theta}_{22} .
    \label{eq:linear_momentum_coords}
\end{split}
\end{equation}
where $\theta_{ij}$ is the absolute angle of the joint from the $x$-axis, and $e_{ij}$, and $l_{ij}$ are the inline center of mass vector and link length as defined in Fig. \ref{fig:fivebl}.

The kinematic velocity loop closure constraint is:
\begin{equation}
    \begin{split}
    \begin{bmatrix}
        -l_{11} \sin(\theta_{11}) \\
        l_{11} \cos(\theta_{11})
    \end{bmatrix}
    \dot{\theta}_{11} & + 
    \begin{bmatrix}
        - l_{12} \sin(\theta_{12}) \\ 
        l_{12} \cos(\theta_{12})
    \end{bmatrix} \dot{\theta}_{12} = \\
    \begin{bmatrix}
        -l_{21} \sin(\theta_{21}) \\
        l_{21} \cos(\theta_{21})
    \end{bmatrix}
    \dot{\theta}_{21} & +
    \begin{bmatrix}
        - l_{22} \sin(\theta_{22}) \\ 
        l_{22} \cos(\theta_{22})
    \end{bmatrix} \dot{\theta}_{22}. 
    \end{split}
    \label{eq:closed_loop_constraint}
\end{equation}

From which \(\dot{\theta}_{22}\) can be derived and substituted into \eqref{eq:linear_momentum_coords}:

\begin{equation}
\begin{split}
    \vect{L} = (m_{11} e_{11} + m_{12} l_{11} + m_{22} e_{22} \frac{l_{11}}{l_{22}})
    \begin{bmatrix}
    -\sin(\theta_{11}) \\ \cos(\theta_{11})
    \end{bmatrix} 
    & \dot{\theta}_{11} \\ 
    + (m_{12} e_{12} + m_{22} e_{22} \frac{l_{12}}{l_{22}})
    \begin{bmatrix}
        -\sin(\theta_{12}) \\
        \cos(\theta_{12})
    \end{bmatrix}
    & \dot{\theta}_{12} \\
    + (m_{21} e_{21} + m_{22} l_{21} - m_{22} e_{22} \frac{l_{21}}{l_{22}})
    \begin{bmatrix}
        - \sin(\theta_{21}) \\
        \cos(\theta_{21})
      \end{bmatrix} 
    & \dot{\theta}_{21} 
    \label{eq:linear_momentum_reduced}
\end{split}
\end{equation}

For a manipulator to be force balanced, the linear momentum must be null (\(\vect{L} = \vect{0}\)), thus one of the non-trivial solutions requires the non-trigonometric coefficients of \eqref{eq:linear_momentum_reduced} to be null:

\begin{align}
    m_{11} e_{11} + m_{12} l_{11} + m_{22} e_{22} \frac{l_{11}}{l_{22}} &  = 0 \label{eq:fivebl_bal_condition_1} \\
    m_{12} \frac{e_{12}}{l_{12}} + m_{22} \frac{e_{22}}{l_{22}} &  = 0 \label{eq:fivebl_bal_condition_2}\\
    m_{21} e_{21} + m_{22} l_{21} \left( 1 - \frac{e_{22}}{l_{22}} \right) &  = 0 .\label{eq:fivebl_bal_condition_3}
\end{align}
These three equations contain 12 variables, of which up to 9 can be selected and the remaining solved for to design a fully force balanced mechanism. 

\subsection{Forbal-2: Planar 2-DOF manipulator}
Starting from the model of Fig. \ref{fig:fivebl}, a force balanced five bar linkage (Forbal-2) was developed.
All the links are of the same length $l$, and the first joints of the two RR chains are placed on the same axis ($l_d=0$).
All links have the same length \(l\) between joints which simplifies the kinematics, maximizes workspace, and ensures that the manipulator does not reach singular configurations within joint limits. Counter masses are only required on 3 links to achieve force balancing for the 5BL, and the total mass is the same or lower compared to placing them on all 4 links according to our previous analysis \cite{VyasEtAl2025Designcomparativeanalysis}. We attach them on Links 11, 21, and 22 through extending the profile behind the direction of the kinematic chain . 

Adjustable counter mass rings are attached to these links at a distance \(l\) behind the joint, imposing additional constraints on \eqref{eq:fivebl_bal_condition_1}-\eqref{eq:fivebl_bal_condition_3}. The proposed design yields a wider workspace, since the missing counter mass on link 12 allows for lesser change of collision between links or with the base. The end-effector is attached to Link 22.

\begin{figure}[!hb]
    \centering
    \includegraphics[trim=0.5cm 1cm 1cm 0.75cm,clip,width=0.9\linewidth]{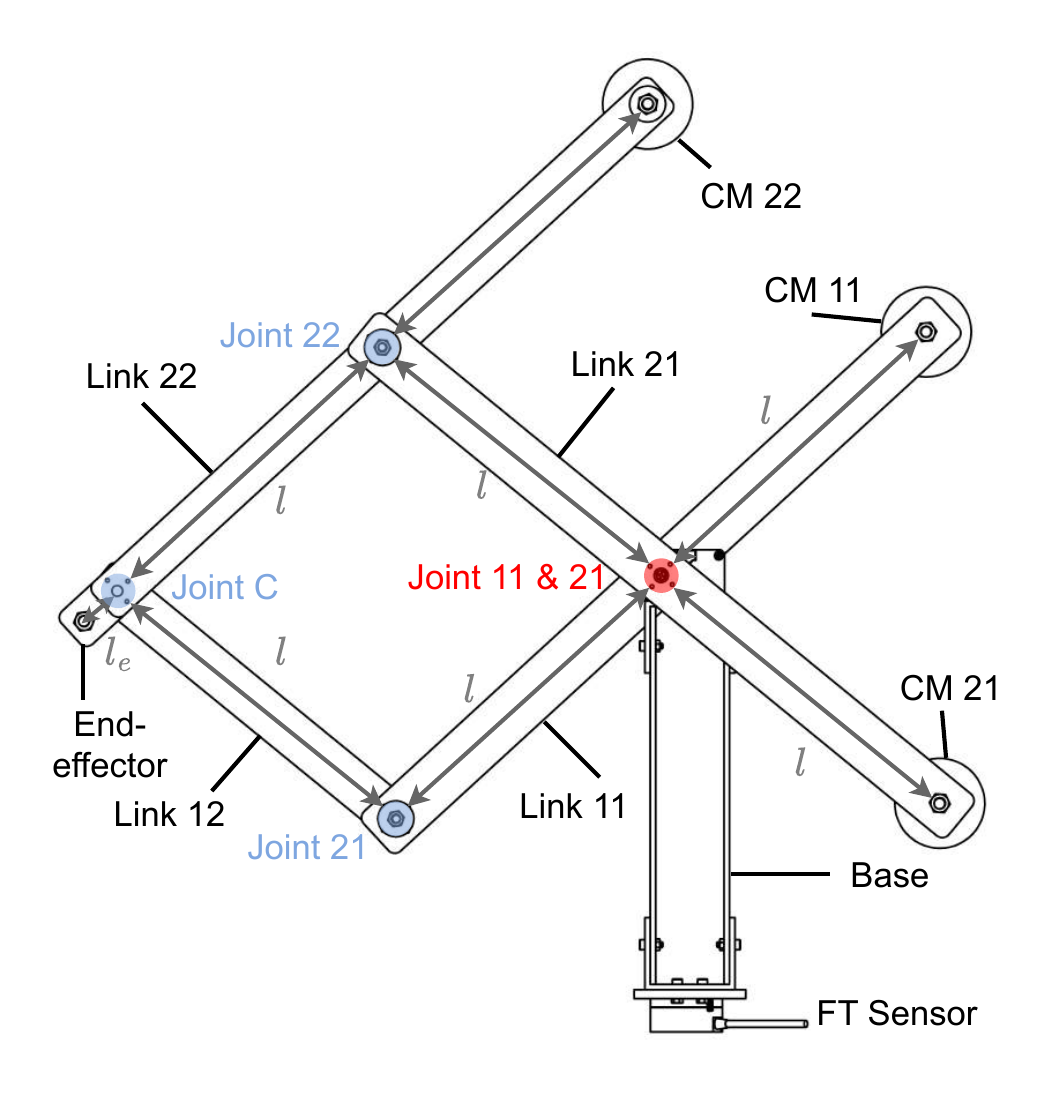}
    \caption{Forbal-2 manipulator design, showing joints (red for actuated and blue for passive), links, and counter masses (CM).}
    \label{fig:forbal2_design}
\end{figure}

To find the counter mass values, it is more convenient to separate the aggregated mass of the link into the profile and counter mass, where subscripts \(p\) and $c$ denote the link profile and the counter mass, respectively: 
\begin{equation}
\begin{split}
    m_{ij} & = m_{ij,p} + m_{ij,c} \\
    m_{ij} e_{ij} & = m_{ij,p} e_{ij,p} + m_{ij,c} e_{ij,c}
\end{split}
\label{eq:mass_separation}
\end{equation}

As there are no counter masses on Link 12, the counter mass for Link 22 \(m_{22,c}\) can be calculated from \eqref{eq:fivebl_bal_condition_2}:
\begin{equation}
        m_{22,c} = -\frac{l_{22}}{e_{22,c}} \left(\frac{m_{22,p} e_{22,p}}{l_{22}} +\frac{m_{12} e_{12}}{l_{12}}\right),
    \label{eq:m22_counter_mass}
\end{equation}
and substituting for \(l_{22}=l_{12}=l\):
\begin{equation}
     m_{22,c} = -\frac{(m_{22,p} e_{22,p} + m_{12} e_{12})}{e_{22,c}}.
     \label{eq:m22_counter_mass_simplified}
\end{equation}

Although link 22 is longer than link 12 (due to its elongation at the back to support the counter mass), most link designs allow for a nearly constant linear density. Hence, $e_{22,p}\approx 0$, while $e_{12}\approx l/2$. Since \(e_{22,c} < 0\) and \(m_{22,p} e_{22,p} < m_{12} e_{12}\), mass $m_{22,c}$ is always positive.

\subsection{Forbal-5: Extension to Spatial 5-DOF manipulator}
\begin{figure}[!hb]
    \centering
    \includegraphics[trim=0.8cm 0.5cm 0.8cm 0.75cm,clip,width=0.9\linewidth]{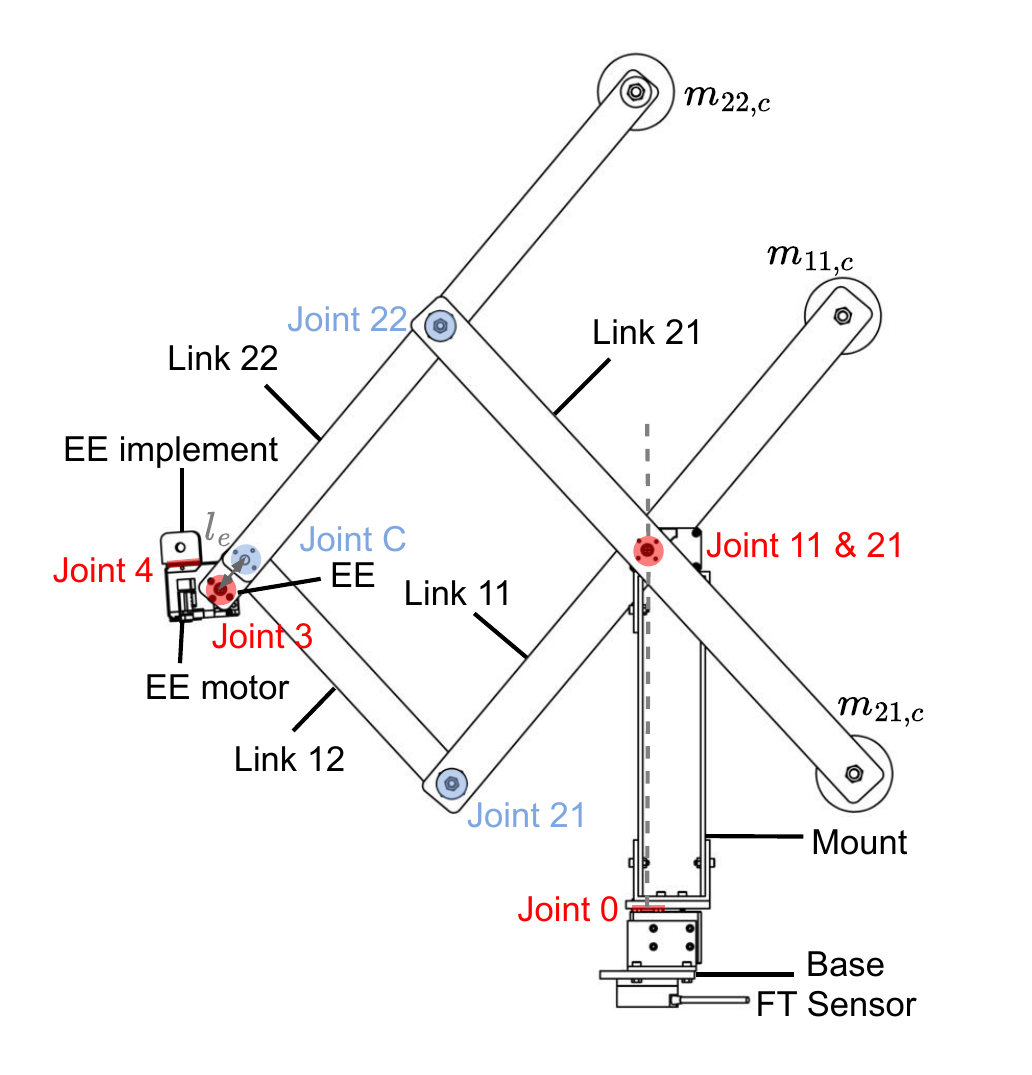}
    \caption{Forbal-5 manipulator design, showing joints (red for actuated and blue for passive), links, and counter masses (CM). 
    }
    \label{fig:forbal5_design}
\end{figure}

The Forbal-2 planar manipulator can be modularly extended to a 5-DOF spatial manipulator through the addition of motors with 3 actuated degrees of freedom, yielding Forbal-5 (Fig. \ref{fig:forbal5_design}). 
The original Forbal-2 manipulator is attached to a mount that can rotate about the inertial \(z\) axis, referred to as Joint 0. The Joint 0 axis is placed directly below the center of Joint 11 and 21 positions, about which the 5BL is force balanced. 

The static end-effector payload of Forbal-2 is replaced with a 2 DOF EE motor. These two additional rotational joints (named Joint 3 and Joint 4) do not intersect each other due to space constraints. Joint 3 is placed on the Forbal-2 end effector position $\vect{p}_e$, and its direction is parallel with Joints 11 and 21. Joint 4 is orthogonal to it, that actuates the EE implement. 

The EE motor and implement together constitute a new end effector payload which should have an inline center of mass, aligned with Link 22 for full force balance, assuming there are no frictional forces acting on the system from the joints and actuator gears. 

The additional degrees of freedom allow for full 3-DOF Cartesian positioning of the end effector as well as a 2-DOF pose. This configuration is suitable for applications on a surface by aligning the implement pose with the surface normal. However, we expect that the center of mass imbalance in the out-of-plane axis increases further, which can be expected to increase the \(x\) axis reaction moment at the base. 
\section{Inverse Kinematic Model}
\label{sec:ik_control}
\subsection{Planar Geometric Controller}
\begin{figure}[!h]
    \centering
    \includegraphics[trim=1cm 1cm 1cm 0.75cm,clip,width=0.9\linewidth]{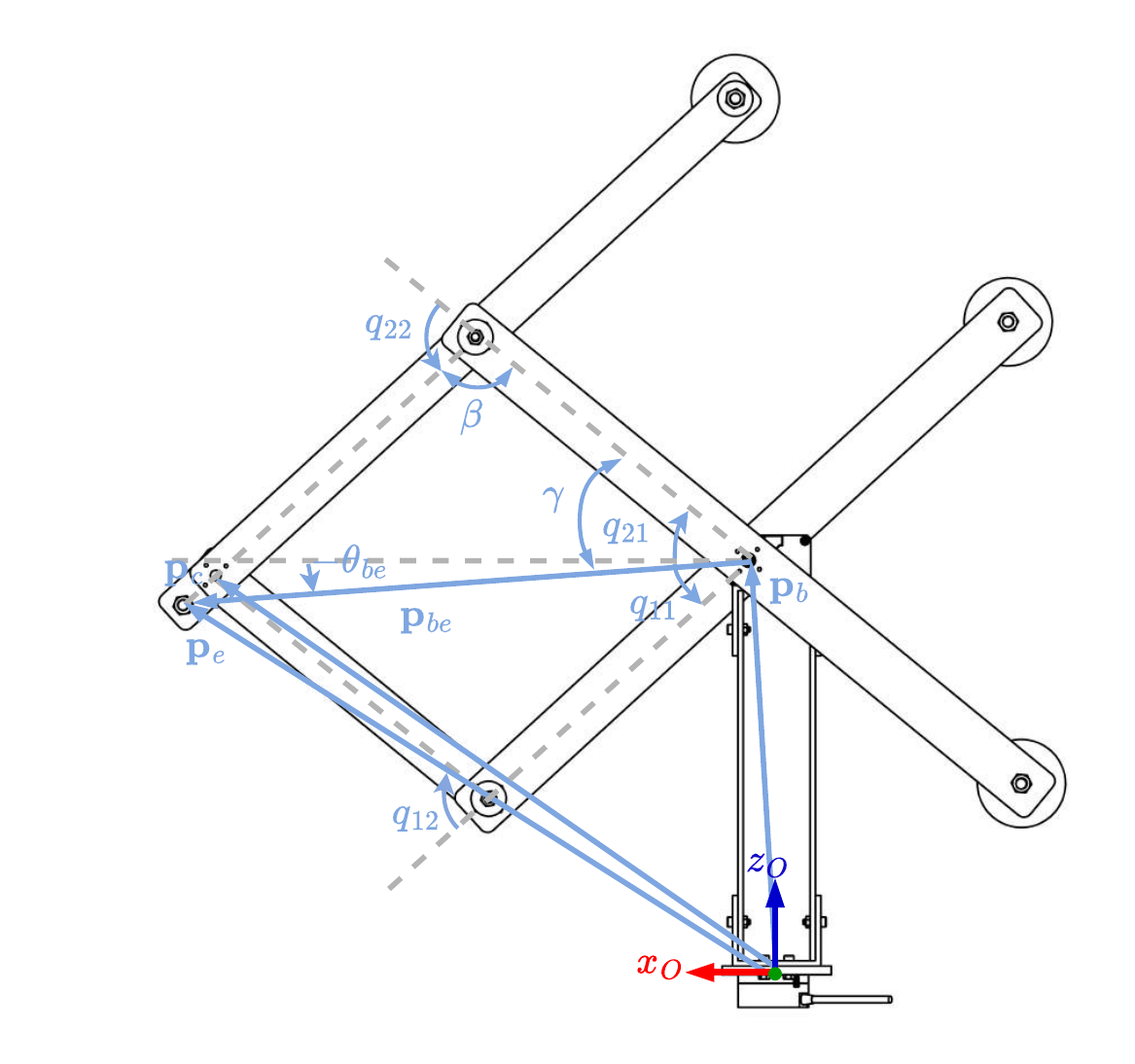}
    \caption{Forbal-2 manipulator. The fixed origin axes are shown as \(x_O\) and \(z_O\), and the joints are aligned to rotate about their y-axes in the xz plane. The cartesian position of the mid point of Joints 11 and 21 in the xz plane is \(\vect{p}_{b}\), and the end-effector xz plane vector is \(\vect{p}_e \). The coordinates for the joints are \(q_{11}\), \(q_{21}\), \(q_{12}\) and \(q_{22}\) as shown. Angles \(\beta\) and \(\gamma\) are used in intermediate computations.}
    \label{fig:forbal2_kinematics}
\end{figure}

The Forbal-2 design closed loop kinematic structure is a rhombus. Therefore, simple trigonometric and geometric calculations can be used to derive the closed-form inverse kinematics. This provides a major advantage of this design as it greatly reduces computation time for the controller, which, along with force balancing, facilitates rapid and precise motions. The kinematics of the manipulator are shown in Fig. \ref{fig:forbal2_kinematics}. A modified Denavit-Hartenberg (DH) convention using \(y\) as the joint axis and aligning joint directions to form a closed loop is used to simplify the inverse kinematic derivation. It is different from that in section \ref{ssec:force_balance_derivation}, hence the symbol \(q\) is used instead of \(\theta\).

Given a fixed frame position vector for the end-effector in the \(xz\) plane \(\vect{p}_e \in \nR{2}\), the inverse kinematics derive the required corresponding joint angle configuration \(q_{11}\) and \(q_{21}\). First, we calculate the vector from the position of the servomotors in the \(xz\) plane, \(\vect{p}_b\) to the end-effector: \(\vect{p}_{be} = \vect{p}_e-\vect{p}_b\). We then derive the end-effector Euclidean distance \(||\vect{p}_{be}|| = \sqrt{p_{be_x}^2+p_{be_z}^2}\) and its angle with respect to the \(x\) axis at \(\vect{p}_b\) as \(\theta_{be} = \tan^{-1}(\frac{p_{be_z}}{p_{be_x}})\).

The angles \(\beta\) and \(\gamma\) are then calculated using the cosine rule:

\begin{equation}
\begin{split}
    \beta &= \cos^{-1}\left(\frac{l^2+(l+l_e)^2-||\vect{p}_{be}||^2}{2 l (l + l_e)}\right) \\
    \gamma &= \cos^{-1}\left(\frac{||\vect{p}_{be}||^2+l^2-(l+l_e)^2}{2l ||\vect{p}_{be}||^2}\right)
\end{split}
\label{eq:forbal2_IK_int}
\end{equation}

The actuated joint angles can then be derived from these internal angles:
\begin{equation}
\begin{split}
    q_{21} &= \gamma + \theta_{be}\\
    q_{11} &= \pi-\beta-q_{21}
\end{split}
\label{eq:forbal2_IK_act}
\end{equation}

The relationship in \eqref{eq:forbal2_IK_act} is always maintained because \(q_{21} + q_{11} > 0\), otherwise the 5BL passes through a singularity. To find the passive joint angles, we can use the geometric constraints of a rhombus with the inner loop angle: \(q_{12} = q_{12} = \pi-\beta\).

\subsection{Forbal-5: Extension of planar IK to 5-DOF pose}
\begin{figure}[!ht]
    \centering
    \includegraphics[trim=1cm 0.9cm 0.9cm 1cm,clip,width=0.8\linewidth]{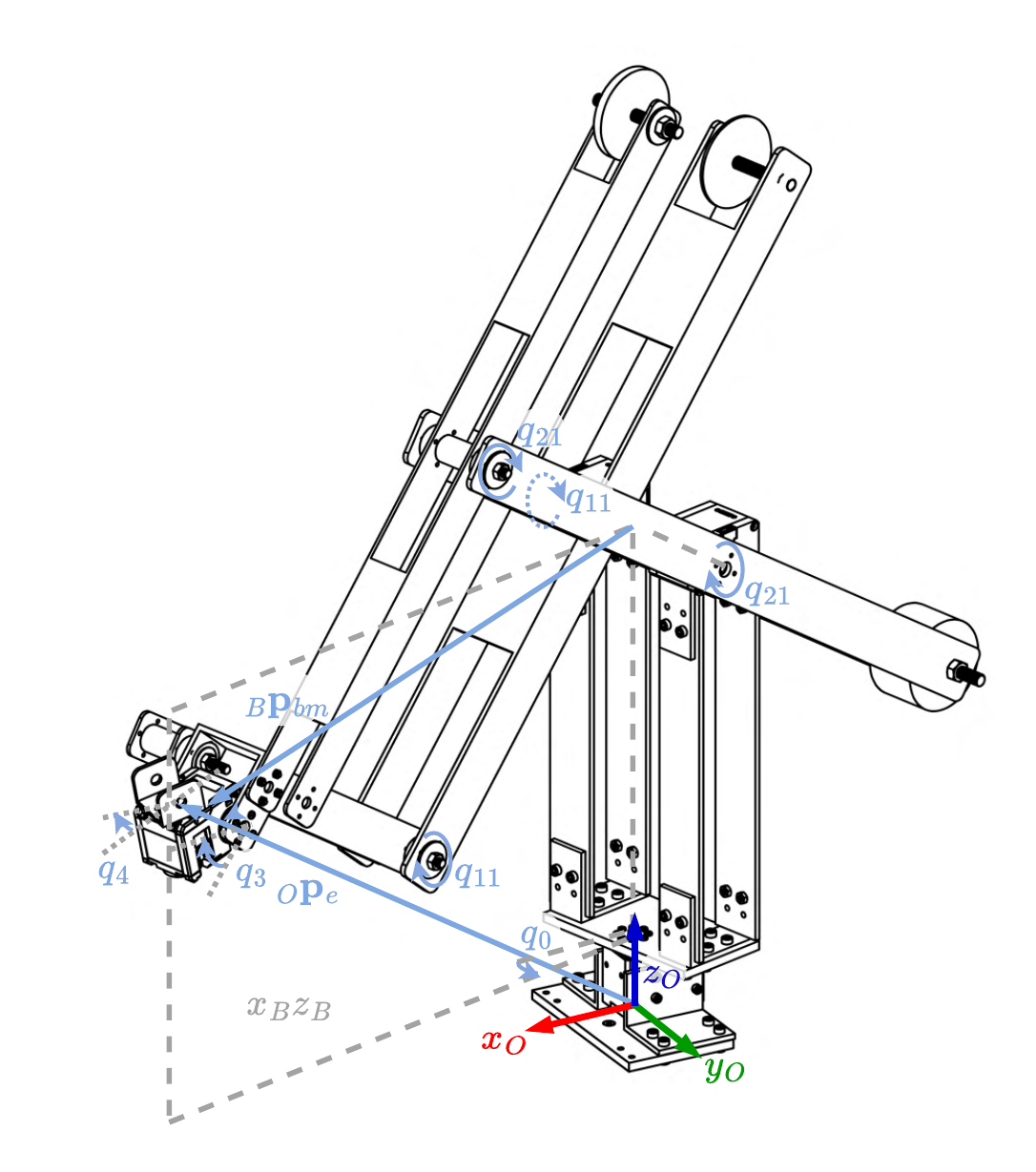}
    \caption{Forbal-5 manipulator. The fixed origin axes are noted as \(x_O\), \(y_O\) and \(z_O\), and the Joint 0 rotated 5BL base frame is shown with the \(x_B z_B\) plane. The end effector position in world frame is \(_O \vect{p}_e\), and \(_B\vect{p}_{bm}\) is the vector from the center of Joints 11 and 21 to the EE motor in the \(x_B z_B\) plane.}
    \label{fig:forbal5_kinematics}
\end{figure}

With the addition of 2-DOF motion, the end-effector implement can be set to a desired 5-DOF pose. We define the EE implement pose as \(\vect{T}_e = \icol{_O p_{e_x} & _O p_{e_y} & _O p_{e_z} & _B p_\beta & _O p_\gamma}\). The first three components are the Cartesian coordinates in the global frame \(_O \vect{p}_e\). 
The next two orient the end-effector; \(_B p_\beta\) is the pitch angle of the EE motor about the 5BL plane \(y_B\) axis orthogonal to \(x_B z_B\) at \(_O \vect{p}_{m}\) (Joint 3 position). The yaw angle \(_O p_\gamma\) is about the global z-axis after the pitch rotation is applied. The final EE implement orientation in the global frame is \(R_y(_B p_\beta)R_z(_O p_\gamma)\). The diagram illustrating the kinematics of Forbal-5 is shown in Fig. \ref{fig:forbal5_kinematics}.

To demonstrate how this end effector definition pose is used, for example, setting \(_O p_\beta=0\),  \(_O p_\gamma=0\), adjusts for the rotation of the end effector from the rotation of \(q_{21}\) and \(q_{22}\) due to the closed loop kinematics, to keep the implement aligned with \(x_O\) and \(y_O\) for all motions of the 5BL.

Since the origins of the reference frames of the two motors are along the axis \(z_0\),
and the center of the EE motor is fixed to the center of the end effector, Joint 4 is directly positioned on the 5BL's \(x_B z_B\) plane, allowing for a simple inverse kinematics formulation that utilizes the previous planar Forbal-2 inverse kinematics with some modifications.

First, we calculate the orientation of the 5BL and joint angle \(q_0\) through a simple trigonometric substitution:

\begin{equation}
    q_0 = \tan^{-1}\left(\frac{_O p_{e_y}}{_O p_{e_x}}\right)
\end{equation}

This defines the orientation of the 5BL frame through the Rotation \(R(q_0) \in SO(3)\), which is used to find the implement Cartesian coordinates in the rotated mount frame:

\begin{equation}
    _B \vect{p}_{e} = R^{-1}(q_0) _O \vect{p}_e
\end{equation}

Note that the y-axis position of the EE implement always lies on the \(x_B z_B\) plane, which rotates about the global z-axis \(z_O\).

We have to adjust the position of the implement about the pitch angle \(p_\beta\) to find the desired coordinates of the EE motor in \(x_B z_B\):

\begin{equation}
    _B \vect{p}_{Om} = {}_B R^{-1}(_B p_\beta) {}_M \vect{p}_{me}
\label{eq:forbal5_IK_pitch}
\end{equation}

where \({}_B R^{-1}(_B p_\beta) \in SO(3) \) is the Y-axis rotation caused by the desired pitch \(_B p_\beta\), and \(_M \vect{p}_{me}\) is the vector from Joint 3 to Joint 4 in the EE motor frame.

To use the Forbal-2 inverse kinematics, we define the target position of the EE motor from the center point of Joint 11 and Joint 21 (the mount axis) in the 5BL frame \(_ B\vect{p}_{bm} = {_B\vect{p}_{Om}} - \vect{p}_{b}\). Using this formulation, \eqref{eq:forbal2_IK_int}-\eqref{eq:forbal2_IK_act} can be used to find \(q_{11}\) and \(q_{21}\). The EE motor joint angle \(q_3\) simply adjusts for the desired pitch \(p_\beta\) about \(y_B\) from \(x_B\) according to the calculated inverse joint angles: \(q_3 = q_{22} - q_{21} + {}_B p_\beta\).

The final calculation is of \(q_4\). Assuming \(_B p_\beta \neq \pm\pi\), which are singular angles for the implement pose, we correct the desired \(_O p_\gamma\) for the mount frame rotation: \(q_4 =  q_0 - {}_B p_\gamma\).

Using this inverse kinematic formulation, we can calculate the joint angles for any desired end effector implement pose within the workspace boundaries. 

\section{Experimental Validation}\label{sec:experiments}
\subsection{Prototype}
The modular Forbal prototype was fabricated using lightweight aluminum link profiles with hollow sections to allow for greater joint angle maneuverability. 
The inertial properties of the manufactured prototype are listed in Table \ref{tab:forbal2_inertial_data}, where the mechanical components for joints (bearings) and other attachments (screws etc.) are included with the respective links, motors are included in the base as they are statically mounted.
The position of the CoMs is defined in each link frame, where the frames are defined according to the modified DH convention.

\begin{table}[!ht]
\centering
\caption{The mass properties for both Forbal-2 and Forbal-5 without balancing CMs. The CoMs are defined in the aforementioned modified DH convention, and total in inertial frame O. The end effector frame is aligned with \(q_{22}\) with origin at the closed loop position \(\vect{p}_c\). }
\begin{tabular}{|l|l|lll|}
\hline
\textbf{Body} & \textbf{Mass (g)} & \multicolumn{3}{l|}{\textbf{CoM (mm)}}                                         \\ \cline{3-5} 
              &                   & \multicolumn{1}{l|}{\textbf{x}} & \multicolumn{1}{l|}{\textbf{y}} & \textbf{z} \\ \hline
Base (Forbal-2)         & 528.1             & \multicolumn{1}{l|}{0.01}       & \multicolumn{1}{l|}{0.00}       & 124.63     \\ \hline
Link 11       & 292.8             & \multicolumn{1}{l|}{-4.89}      & \multicolumn{1}{l|}{-0.01}      & 2.30       \\ \hline
Link 12       & 95.5              & \multicolumn{1}{l|}{122.57}     & \multicolumn{1}{l|}{17.30}      & 2.11       \\ \hline
Link 21       & 291.3             & \multicolumn{1}{l|}{-11.96}     & \multicolumn{1}{l|}{-0.42}      & 1.81       \\ \hline
Link 22       & 153.7             & \multicolumn{1}{l|}{-22.75}     & \multicolumn{1}{l|}{-15.13}     & 1.89       \\ \hline
EE (Forbal-2) & 43.7              & \multicolumn{1}{l|}{10.99}      & \multicolumn{1}{l|}{0.85}       & 0.00       \\ \hline
EE (Forbal-5) & 27.1              & \multicolumn{1}{l|}{2.44}       & \multicolumn{1}{l|}{1.38}       & 0.00       \\ \hline
EE motor      & 102.0             & \multicolumn{1}{l|}{12.0}       & \multicolumn{1}{l|}{4.90}       & 0.01       \\ \hline
EE implement  & 3.02              & \multicolumn{1}{l|}{0.00}       & \multicolumn{1}{l|}{0.00}       & 6.37       \\ \hline
\end{tabular}
\label{tab:forbal2_inertial_data}
\end{table}

Link 12 is kept short with no backwards extension, to facilitate workspace as keeping it longer would greatly reduce the motion of the manipulator. We find a further advantage in that this does not significantly increase the total mass after balancing with counter masses. 

The force balance condition between Link 12 and 22 in \eqref{eq:fivebl_bal_condition_2} is 
can only be fulfilled when \(e_{12} = e_{22} = 0\), or in the case of equal length, \(m_{12} e_{12} = -m_{22} e_{22}\), i.e. the counter mass on one of the links balances the other. Note that the end effector mass is added to Link 22, whereas Link 12 has no additional loads.

If we use the same profile for Link 12 and 22, the force balance requires \(e_{12} = e_{22} = 0\), which makes the counter mass calculation straightforward:
\begin{equation}
    m_{ij,c} = \frac{-m_{ij,p} e_{ij,p}}{e_{ij,c}}
    \label{eq:counter_mass_same_prof}
\end{equation}

In both cases, \(m_{12}\), \(m_{22}\), \(e_{12}\), and \(e_{22}\) with the counter masses for force balance are known, and \eqref{eq:mass_separation} can be substituted for the remaining links in \eqref{eq:fivebl_bal_condition_1} and \eqref{eq:fivebl_bal_condition_3} to find counter mass values \(m_{11,c}\) and \(m_{21,c}\). 

We also performed an evaluation of the changes in the counter masses due to the reduction in length of Link 12 to increase the manipulator workspace. From this point forward, let us call \textit{Link 12 short} the solution without the counter mass on Link 12 (Figure \ref{fig:forbal2_design}), and \textit{Link 12 extended} the version with the counter mass on Link 12 (and a smaller workspace due to possible interaction with the ground on which the manipulator is installed). 
A comparison of the counter mass and total mass values the link are shown for the two aforementioned cases in Table \ref{tab:forbal2_link12_comparison}: 1) Link 12 with the selected short profile without counter masses, and 2) the same profile as Link 22 with counter masses.

\begin{table}[!ht]
\centering
\caption{Forbal-2: comparison of counter masses Link 12 with short and extended profile (same as Link 22), and the resulting total manipulator mass after balancing.}
\begin{tabular}{|l|l|l|}
\hline
\textbf{Mass (g)}      & \textbf{Link 12 short} & \textbf{Link 12 extended} \\ \hline
\(m_{11,c}\)  & 29.8         & 164.0             \\ \hline
\(m_{12,c}\)  & -            & 0.000             \\ \hline
\(m_{21,c}\)  & 325.6        & 173.5             \\ \hline
\(m_{22,c}\)  & 87.1        & 11.1              \\ \hline
Total without CM & 1405.1    & 1463.1             \\ \hline
Total with CM & 1847.6       & 1811.7            \\ \hline
\end{tabular}
\label{tab:forbal2_link12_comparison}
\end{table}

As demonstrated in Table \ref{tab:forbal2_link12_comparison}, no additional counter mass is required for Link 12 extended as the extended profile balances the link. Using the Link 12 short profile only increases the mass by around 36g after balancing, but greatly increases the workspace by enabling a wider joint limit for \(q_{12}\) without collision with the base. This was also observed for abstract high-level design in earlier studies. However, Link 12 extended presents a higher counter mass stacked on Link 21. With the short profile for Link 12, the center of mass is at \(y=9.89\)mm at the default configuration, while for the longer link profile it is at \(y=1.52\)mm.  This will increase the reaction moment in the \(y_0\)-axis due to the increased center of mass offset from the ideal of \(y=0\)m. 

The counter masses are implemented through two types of steel rings mounted on a filleted M6 steel shaft and nuts: 1) M6, 40mm diameter, thickness (mass per ring 13.2g), and 2) M8, 32mm, thickness 2mm (mass per ring 11.4g). Using rings over fixed mass segments allowed us to adjust the counter masses for different payloads at the end-effector, and also compensate manually for differences between the model and actual prototype mass.

We also note that the designed links are not perfectly inline due to the slight asymmetries in the profile shape and additional loads applied. However, in practice we found that due to static friction from the joints and motor, the configuration remains easily force balanced for any inline load without any gravity compensation torque.

The passive revolute joints are made out of Teflon and M5 bolt/nut with spacers for fastening. The selection of this material allowed for easy assembly and disassembly with low dynamic friction over the operating velocity range.

Joints 11 and 21 are actuated through Dynamixel XC430-W240T servomotors. They have a maximum stall torque of 1.9\(\Nm\) and a positional encoder with \(0.089^\circ\) resolution. This servomotor provides enough torque to actuate the manipulator at reasonably fast joint velocities of 7 rad/s. 

For Forbal-5, we selected the high torque Dynamixel XM430-W350-T motor for Joint 0, and a 2-axis Dynamixel 2XL430-W250-T motor to implement Joints 3 and 4 in the EE motor. Due to the placement of joints in the motor, once Joint 3 rotates the 5BL is no longer inline and theoretically the masses are only partially balancing the payload.  Here we found again that the friction from the passive joints and gears in the motor facilitate force balancing. 

The manipulator was balanced with calculating the counter masses based on the CAD model, and then adjusted with trial and error to fit the actual prototype. This accounted for the modeling inaccuracies, as it was neither possible to accurately determine the link mass properties nor to adjust counter masses to the gram. The final counter mass values compared to theoretical are shown in Table \ref{tab:actual_countermasses}. 

\begin{table}[!ht]
\centering
\caption{Experimentally applied counter masses for force balancing of Forbal-2 and Forbal-5. Quantities are in grams.}
\begin{tabular}{|l|rl|rl|}
\hline
\textbf{Link} & \multicolumn{2}{c|}{\textbf{Forbal-2}}                      & \multicolumn{2}{c|}{\textbf{Forbal-5}}                      \\ \cline{2-5} 
              & \multicolumn{1}{l|}{\textit{Theoretical}} & \textit{Actual} & \multicolumn{1}{l|}{\textit{Theoretical}} & \textit{Actual} \\ \hline
11            & \multicolumn{1}{r|}{29.8}                 & 27.0            & \multicolumn{1}{r|}{29.8}                 & 29.8            \\ \hline
21            & \multicolumn{1}{r|}{325.6}                & 331.6           & \multicolumn{1}{r|}{515.5}                & 535.8           \\ \hline
22            & \multicolumn{1}{r|}{87.1}                 & 58.1            & \multicolumn{1}{r|}{190.1}                & 189.0           \\ \hline
Total with CM & \multicolumn{1}{l|}{1847.6}               & 1821.8          & \multicolumn{1}{l|}{2421.5}               & 2440.7          \\ \hline
\end{tabular}
\label{tab:actual_countermasses}
\end{table} 

\begin{figure}[!ht]
    \centering
    \includegraphics[width=\linewidth]{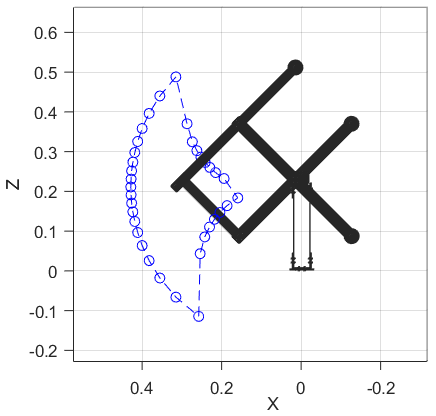}
    \caption{Workspace of the Forbal-2 manipulator.}
    \label{fig:forbal-2_workspace}
\end{figure}

The joint limits to prevent self-collision and avoid singularities are shown in Table \ref{tab:joint_limits_forbal-2}. 
A workspace analysis was conducted with constant velocity trajectories radially spaced at 10\(^\circ\) from a default configuration of \(q_{11}=q_{21}=45^\circ\) rad, determining a workspace of \(0.081\msq\) with a maximum reach of 0.605m. The workspace is depicted in Fig. \ref{fig:forbal-2_workspace}. The workspace shape resembles an axe, with center at \(\vect{p}_e=\icol{0.3 & 0.18}\)m from which the maximum span can be reached. The workspace of Forbal-5 is a toroid with the cross section the Forbal-2 workspace adjusted for the EE motor offset, giving it a volume of 0.102\(\mcb\).

\begin{table}[!ht]
\small\sf\centering
\caption{Joint limits for Forbal-2}
\begin{tabular}{|l|l|l|l|l|}
\hline
\textbf{Joint}     & \(q_{11}\) & \(q_{12}\) & \(q_{21}\) & \(q_{22}\) \\ \hline
\textbf{Min (rad)} & -1.25      & -2.9       & -1.5       & 0.3        \\ \hline
\textbf{Max (rad)} & 1.45       & 2.9        & 1.28       & 2.9        \\ \hline
\end{tabular}
\label{tab:joint_limits_forbal-2}
\end{table}

\subsection{Test setup}
Both the balanced and unbalanced configurations of Forbal-2 and Forbal-5 were tested with various trajectories to validate and measure the impact of force balancing on reaction forces and moments, joint torques, and end-effector position error. 
For each trajectory, we present experiments with and without balancing counter masses, noted in the results as balanced and unbalanced configurations.

The manipulator was mounted on a wooden test base, which is clamped to the test bench. 
A ROBOTOUS 100N Force/Torque sensor was attached coincident with the base reference frame to measure reaction forces and torques. The sensor measurements were sampled at 500Hz, and low-pass filtered at 50Hz to remove high-frequency noise. 

Due to small misalignment with the sensor and the position of the manipulator CoM, the sensor was calibrated with a manually set constant offset, first by setting the no-load z-force to 0, and then by mounting the manipulator and adjusting the forces and torques in the remaining axes at the nominal pose to 0.

The Dynamixel servomotors have an internal microcontroller with closed loop feedback, which is set through the U2D2 communication bridge at a maximum frequency of 62.5 Hz. The position commands and sensor readings, such as joint angle and current (to estimate the torque), were sent through the U2D2 communication bridge using the ROS2 control library. For external validation and analysis of the manipulator precision, we used a 3.1 Megapixel Genie Nano 5G M2050 Mono Camera at 160 frames per second to localize the end-effector through April tags and measure the position error from the reference trajectory. 

The high-level position controller took as input a set of time spaced end-effector positions / pose waypoints. It interpolated a smooth cubic Hermite spline path in position or joint space, sampled them with a discrete time step set to 0.01s, and scaled the motion profile according to a trapezoidal speed law determined through acceleration and deceleration times. This resulted in trajectories that were continuous in velocity, and removed initial jerk in the motion. However, the timing of waypoints shifted from the original based on the acceleration time as the trapezoidal speed law was applied after the waypoint splines were generated.

The inverse kinematic equations converted the reference position space trajectory to a joint space trajectory and commanded to ROS2 control. Dynamixel servomotors contain a low-level closed-loop motor controller with tunable PID motor gains, which we set as shown in Table \ref{tab:motor_gains}. We tuned them manually with PD gains to result in the fastest stable step response without introducing instability in the form of vibrations. The Integral gain was set to 0 to improve the accuracy in current measurements as we observed that it resulted in a constant current increase over time.

\begin{table}[!hb]
\centering
\caption{Dynamixel motor controller gain register values for Forbal-2 and Forbal-5 experiments.}
\begin{tabular}{|l|l|l|l|lll|}
\hline
Joint    & Gain    & \(q_{11}\)                 & \(q_{21}\)                 & \multicolumn{1}{l|}{\(q_0\)} & \multicolumn{1}{l|}{\(q_3\)} & \(q_4\) \\ \hline
Forbal-2 & \(Kp\)  & \multicolumn{1}{r|}{10000} & \multicolumn{1}{r|}{14000} & -                            & -                            & -       \\ \cline{2-4}
         & \(K_d\) & 12000                      & 16000                      & -                            & -                            & -       \\ \hline
Forbal-5 & \(Kp\)  & \multicolumn{1}{r|}{12500} & \multicolumn{1}{r|}{15000} & \multicolumn{1}{l|}{1200}    & \multicolumn{1}{l|}{2000}    & 2000    \\ \cline{2-7} 
         & \(K_d\) & 16000                      & 18000                      & \multicolumn{1}{l|}{2400}    & \multicolumn{1}{l|}{4000}    & 4000    \\ \hline
\end{tabular}
\label{tab:motor_gains}
\end{table}

\subsection{Forbal-2 Experiments}
For each trajectory, we present the waypoints, the end-effector and joint angle profiles, and the measured reaction forces/moments and joint torques averaged across 5 experiments. The acceleration times in the trapezoidal speed law was set to 0.5s and discretization step of 0.01s, which is higher than the Dynamixel communication frequency. For all the trajectories, we present the reaction forces/moments and joint torques averaged for 5 experiments each in balanced and unbalanced configurations.

\subsubsection{Trajectory 1}
\begin{figure}[!ht]
    \centering
    \begin{subfigure}[b]{\linewidth}
         \includegraphics[trim=3.5cm 11.25cm 4cm 11cm,clip,width=\linewidth]{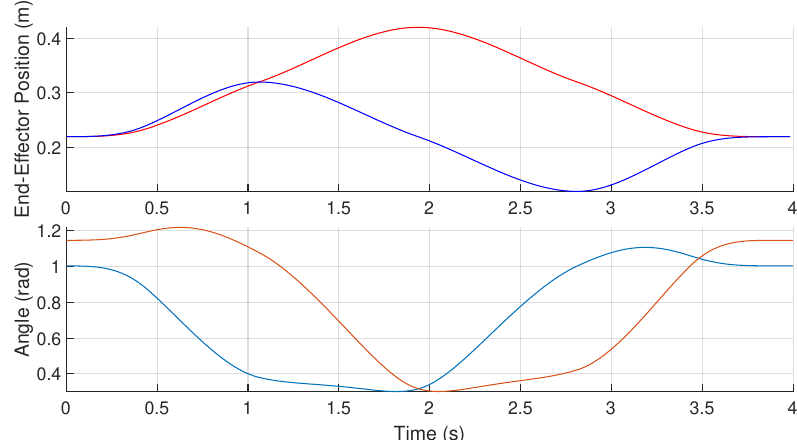}
        \caption{Forbal-2 Trajectory 1 motion profile. }
         \label{fig:forbal2_traj1}
     \end{subfigure}
    \begin{subfigure}[b]{\linewidth}
         \centering
        \includegraphics[trim=3.5cm 7.5cm 4cm 8cm,clip,width=\linewidth]{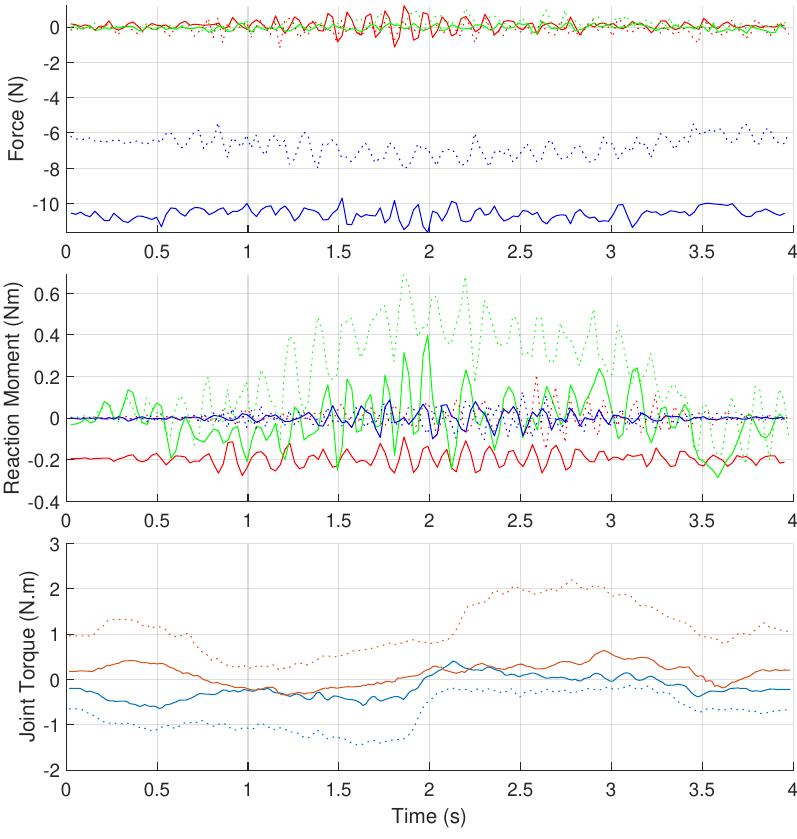}
        \caption{Trajectory 1 results averaged across 5 experiments. The solid line is balanced while dotted is unbalanced.}
    \label{fig:forbal2_traj1_results}
     \end{subfigure}
    \caption{Forbal-2: Trajectory 1. Red (x), green (y) and blue (z) are positions and moments in the inertial frame, while light blue (\(q_{11}\)) and brown (\(q_{22}\)) are joint angles/torques.}
    \label{fig:forbal2_traj1_full}
\end{figure}

has 5 waypoints traversing a closed smooth diamond shape between \(\vect{p}_e = \icol{0.22m & 0.22m}\), \(\vect{p}_e = \icol{0.32m & 0.32m}\), \(\vect{p}_e = \icol{0.42m & 0.32m}\) and \(\vect{p}_e = \icol{0.22m & 0.22m}\), and back to the start in 4 seconds. The motion profile and experimental results are shown in Fig. \ref{fig:forbal2_traj1_full}.

\subsubsection{Trajectory 2}
\begin{figure}[!ht]
    \centering
    \begin{subfigure}[b]{\linewidth}
          \includegraphics[trim=3.5cm 11.25cm 4cm 11cm,clip,width=\linewidth]{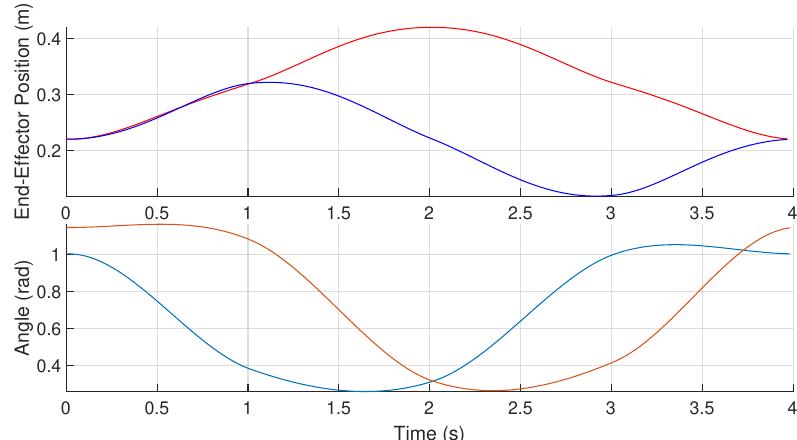}
        \caption{Forbal-2 Trajectory 2 motion profile.}
    \label{fig:forbal2_traj2}
     \end{subfigure}
   \begin{subfigure}[b]{\linewidth}
        \centering
        \includegraphics[trim=3.5cm 7.5cm 4cm 8cm,clip,width=\linewidth]{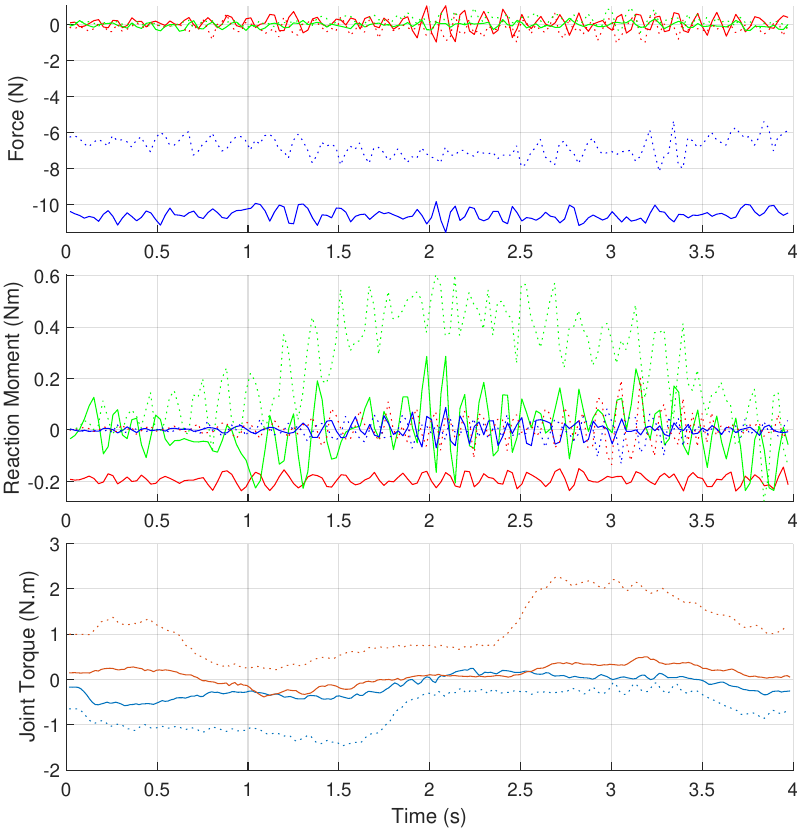}
        \caption{Trajectory 2 results averaged across 5 experiments. The solid line is balanced while dotted is unbalanced.}
        \label{fig:forbal2_traj2_results}
     \end{subfigure}
    \caption{Forbal-2: Trajectory 2. Red (x), green (y) and blue (z) are positions and moments in the inertial frame, while light blue (\(q_{11}\)) and brown (\(q_{22}\)) are joint angles/torques.}
    \label{fig:forbal2_traj2_full}
\end{figure}

 traverses the same waypoints as Trajectory 1, but interpolates the spline trajectory between these waypoints in the joint angle space, resulting in minimal joint velocities and a smoother joint angle profile. The motion profile and experimental results are shown in Fig. \ref{fig:forbal2_traj2_full}.

\subsubsection{Trajectory 3}
\begin{figure}[!ht]
    \centering
    \begin{subfigure}[b]{\linewidth}
        \includegraphics[trim=3.5cm 11.25cm 4cm 11cm,clip,width=\linewidth]{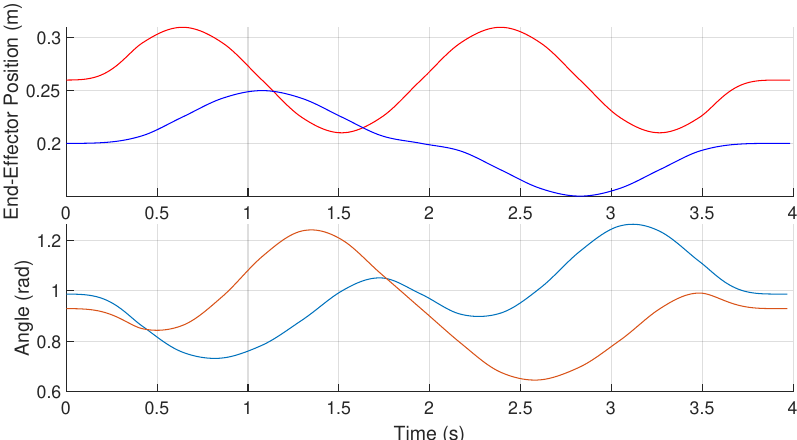}
        \caption{Forbal-2 Trajectory 3 motion profile.}
        \label{fig:forbal2_traj3}
     \end{subfigure}
    \begin{subfigure}[b]{\linewidth}
        \centering
        \centering
        \includegraphics[trim=3.5cm 7.5cm 4cm 8cm,clip,width=\linewidth]{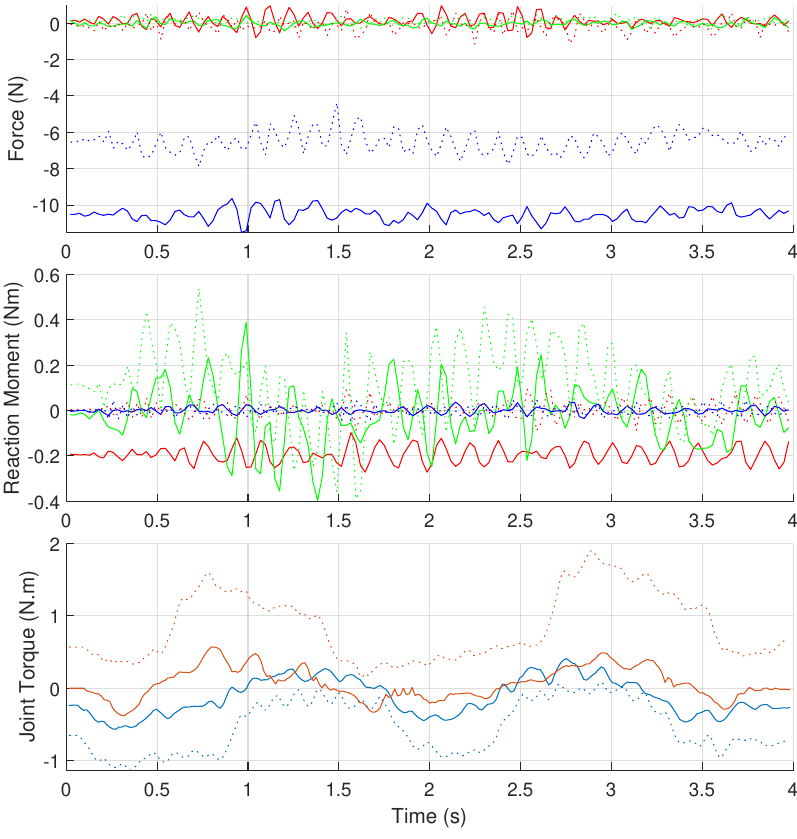}
        \caption{Trajectory 3 results averaged across 5 experiments. The solid line is balanced while dotted is unbalanced.}
        \label{fig:forbal2_traj3_results}
     \end{subfigure}
    \caption{Forbal-2: Trajectory 3. Red (x), green (y) and blue (z) are positions and moments in the inertial frame, while light blue (\(q_{11}\)) and brown (\(q_{22}\)) are joint angles/torques.}
    \label{fig:forbal2_traj3_full}
\end{figure}

measures fast directional changes in the end-effector position through drawing a figure of 8, set with 17 waypoints spaced at 0.25s increments from \(t=0\) to \(t=4\) that sample on 2 circles with a radius of 0.025m. The first circle is traversed from \(t=0\) to \(t=2\), centered around \(\icol{x & z} = \icol{0.235 & 0.225}\). The second circle is traversed from \(t=2\) to \(t=4\) centered at \(\icol{x & z} = \icol{0.285 & 0.175}\). The motion profile and experimental results are shown in Fig. \ref{fig:forbal2_traj3_full}.

\subsubsection{Trajectory 4}
\begin{figure}[!ht]
    \centering
    \begin{subfigure}[b]{\linewidth}
        \includegraphics[trim=3.5cm 11.25cm 4cm 11cm,clip,width=\linewidth]{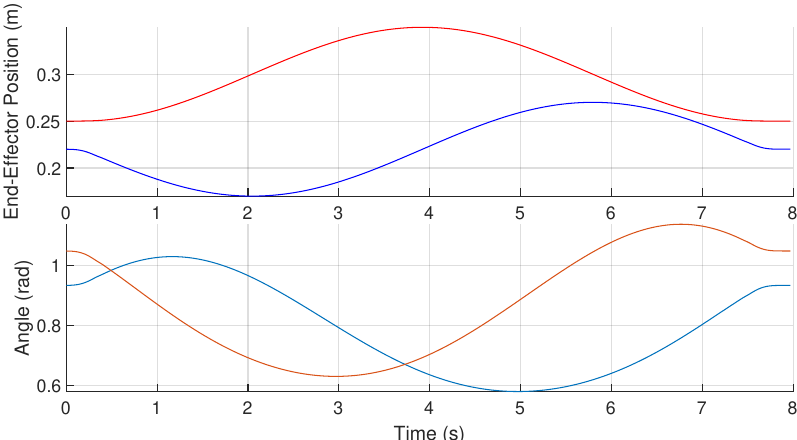}
        \caption{Forbal-2 Trajectory 4 motion profile.}
        \label{fig:forbal2_traj4}
    \end{subfigure}
    \begin{subfigure}[b]{\linewidth}
        \centering
        \includegraphics[trim=3.5cm 7.5cm 4cm 8cm,clip,width=\linewidth]{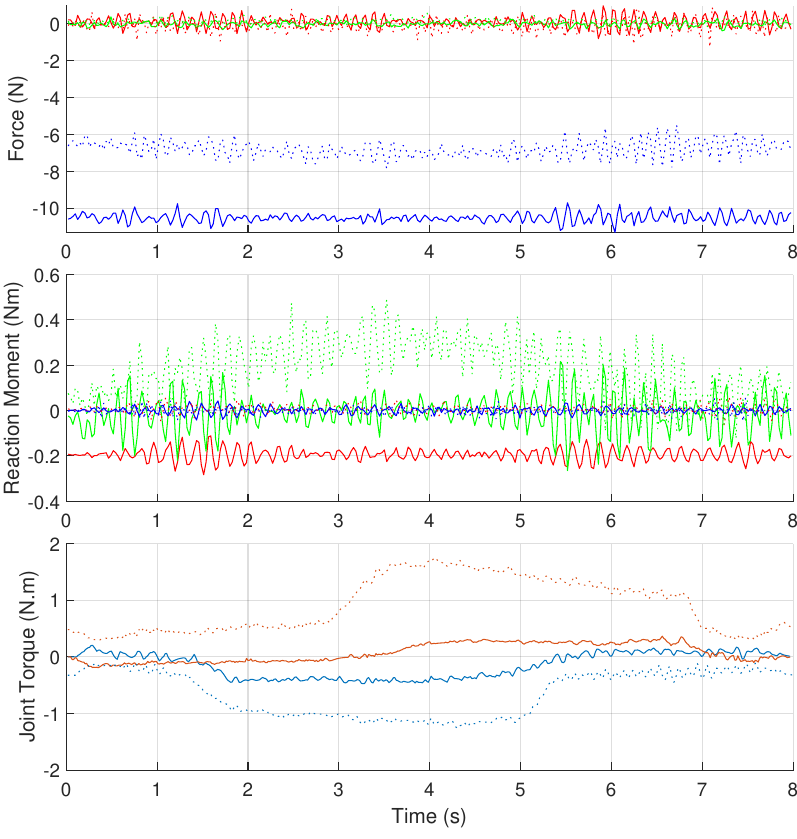}
        \caption{Trajectory 4 results averaged across 5 experiments. The solid line is balanced while dotted is unbalanced.}
        \label{fig:forbal2_traj4_results}
     \end{subfigure}
    \caption{Forbal-2: Trajectory 4. Red (x), green (y) and blue (z) are positions and moments in the inertial frame, while light blue (\(q_{11}\)) and brown (\(q_{22}\)) are joint angles/torques.}
    \label{fig:forbal2_traj4_full}
\end{figure}

slowly traverses a circle with radius 0.05m centered at \(\icol{x_O & z_O} = \icol{0.25 & 0.22}\) in 8 seconds. The circle is sampled with 41 waypoints incremented at 0.2s, where the spline interpolation between these waypoints in end effector positions traces the circle. The motion profile and experimental results are shown in Fig. \ref{fig:forbal2_traj4_full}.

\subsection{Discussion of Forbal-2 experimental results}
We first visually observed the impact of balancing with counter masses, through moving the manipulator links freely without the motor power connected. This is evidenced in Figures \ref{fig:forbal-2} and \ref{fig:forbal-5} where the joint angles are maintained purely through the force balancing properties of both manipulator variants.

The trajectories were selected to determine the impact of force balancing for both fast and slow accelerations. As the acceleration increases, the increase in the inertia from the balancing counter masses causes a higher reaction moment that overpowers the reduction of gravity compensation forces and torques. However, for our experimented trajectories, we find a consistent improvement in the planar Y reaction moment and joint torques for the balanced manipulator.

The mean absolute Y (planar) reaction moment is reduced by average of 41-66\%, and the maximum absolute Y-moment is reduced by up to 53\% for the balanced configuration. Average absolute joint torques are reduced by an average of 54-84\% and the absolute maximum joint torque up to 79\%. For faster trajectories (1 and 3), the improvements from balancing are lesser whereas interpolating smooth trajectories in joint space (2) or a slower trajectory (4) shows stronger improvement.

As expected, the addition of the balancing counter masses exerts an X-axis reaction moment of -0.2 N.m, as the majority of them are placed on Links 21 and 22, which are on one side of the manipulator. The average reaction force in the Z (gravity) axis is also increased for the balanced manipulator due to the additional weight of the counter masses.  

Contrary to expectation, we did not observe a significant difference in the X (in-plane gravity orthogonal) reaction force between the balanced and unbalanced configurations. This can be explained by the fact that Links 11, 21 and 22 are almost fully symmetric in all axes, which significantly contributes to force balance. The balancing counter masses simply adjust for any remaining unbalancing effects from Link 12 and additional loads such as the end effector which in this case is minimal compared to an asymmetric link design.

When comparing the position error results from applying the forward kinematics to the servomotor reference and measured joint angles, we find a consistent but non-significant difference between balanced and unbalanced configurations in the order of 0.1mm. Furthermore, using encoder measurements assumes perfect rigidity of the manipulator links. To further validate the precision of the balanced manipulator, we rely on the external camera tracking of the end-effector.

Due to lighting and motion variations, some experiments did not detect the April Tag consistently for the full trajectory and had to be discarded when finding the precision according to the camera localization. The position error metrics as measured through the camera across all considered experiments are shown in Table \ref{tab:forbal2_camera_results}. 

\begin{table}[!ht]
\centering
\caption{Forbal-2 camera tracked position error. Quantities are in mm, categorized by the trajectory, configuration (config.), experiments considered (Exp.). The mean, standard deviation (std), minimum, and maximum are for all considered experiments.}
\begin{tabular}{|l|l|l|l|l|l|l|}
\hline
\textbf{Trajectory} & \textbf{Config.} & \textbf{Exp.} & \textbf{Mean} & \textbf{Std} & \textbf{Min} & \textbf{Max}  \\ \hline
1          & Bal.   & 5       & 6.6  & 3.3 & 0.6 & 13.4 \\ \cline{2-7} 
           & Unbal.   & 3       & 8.2  & 3.2 & 3.1 & 14.7 \\ \hline
2          & Bal.   & 5       & 6.3  & 2.7 & 1.2 & 11.5 \\ \cline{2-7} 
           & Unbal.   & 5       & 8.6  & 3.3 & 2.8 & 14.8 \\ \hline
3          & Bal.   & 5       & 3.6  & 1.3 & 0.1 & 7.0  \\ \cline{2-7} 
           & Unbal.   & 4       & 6.1  & 2.0 & 0.2 & 11.1 \\ \hline
4          & Bal.   & 5       & 2.5  & 0.7 & 0.3 & 5.2  \\ \cline{2-7} 
           & Unbal.   & 5       & 5.8  & 2.4 & 0.1 & 11.6 \\ \hline
\end{tabular}
\label{tab:forbal2_camera_results}
\end{table}

From the camera results, we see an improvement of 20-56\% in the mean position error for the balanced manipulator over the unbalanced manipulator. This is a conservative estimate, as the detection of April tags by the camera was not perfect, with occasional frames where the tag was not detected, and noise in the detected position. To solve this issue, we interpolated the tag position for missing frames and smoothed the trajectory with Savitsky-Golay filtering. A more accurate motion capture system could give a more accurate and confident estimation of the improvement to precision from balancing the manipulator.

\subsection{Forbal-5 Experiments}
The trajectories for Forbal-5 were selected to measure the changes in force balancing behavior with an increase in load from the end-effector motor, as well as demonstrate a practical engineering application by setting a 5DOF end-effector pose which maintains a constant angle for the end-effector implement while tracking a 3D spatial trajectory. 

\subsubsection{Trajectory 1} 
\begin{figure}[!ht]
    \centering
    \includegraphics[trim=3.5cm 7.5cm 4cm 8cm,clip,width=\linewidth]{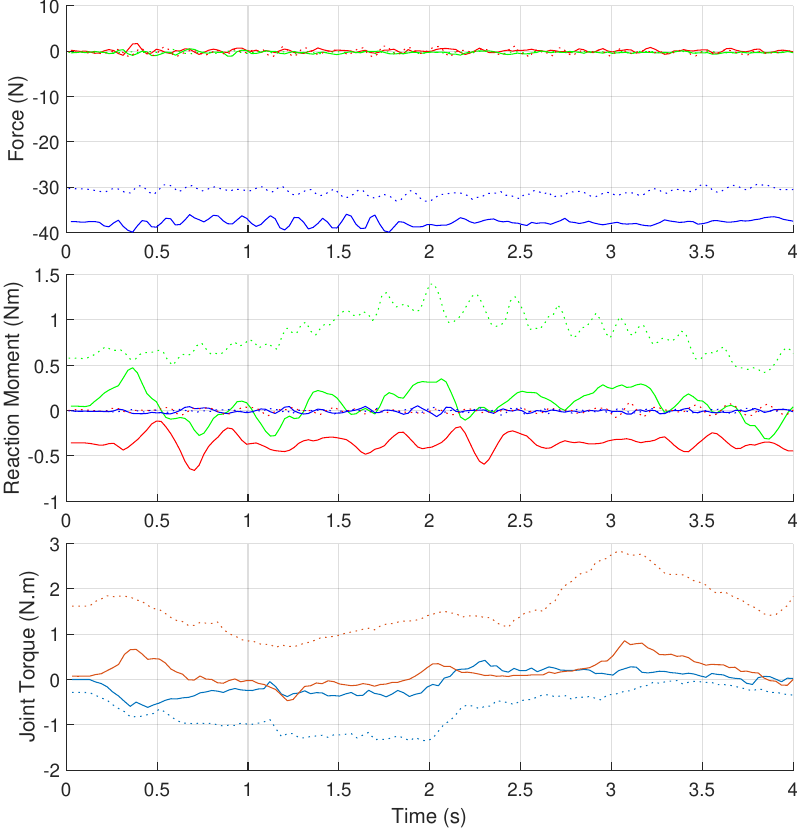}
    \caption{Forbal-5 Trajectory 1 results. The reaction forces and moments are shown as red (x), green (y) and blue (z), while joint torques are light blue (\(q_{11}\)) and brown (\(q_{22}\)). Solid line is balanced while dotted is unbalanced.}
    \label{fig:forbal5_traj1_results}
\end{figure}

implemented on Forbal-5 tracked the same planar waypoints as Trajectory 1 of Forbal-2 with \(p_y = 0\), while also maintaining a steady \(_B p_\beta = 0\) and \(_O p_\gamma = 0\) for the whole trajectory, which kept the end-effector implement level with the xy plane. For brevity's sake, we omit the trajectory profile as it is the same as Figure \ref{fig:forbal2_traj1}. The measured reaction forces/moments as well as the joint torques are shown in Figure \ref{fig:forbal5_traj1_results}. The joint torque for \(q_0\) was negligible as the motion is planar, and the absolute angle error for \(_B p_\beta\) and \(_O p_\gamma\) was below \(10^{-3}\) rad at all times.

\subsubsection{Trajectory 2}
on Forbal-5 was the same as Trajectory 4 of Forbal-2, but also maintained \(_B p_\beta = 0\) and \(_O p_\gamma = -1.570796\) for the whole trajectory, keeping the end-effector implement level with the xy plane. For brevity's sake, we also omit the trajectory profile as it is the same as Figure \ref{fig:forbal2_traj4}. 

\begin{figure}[!ht]
    \centering
    \includegraphics[trim=3.5cm 7.5cm 4cm 8cm,clip,width=\linewidth]{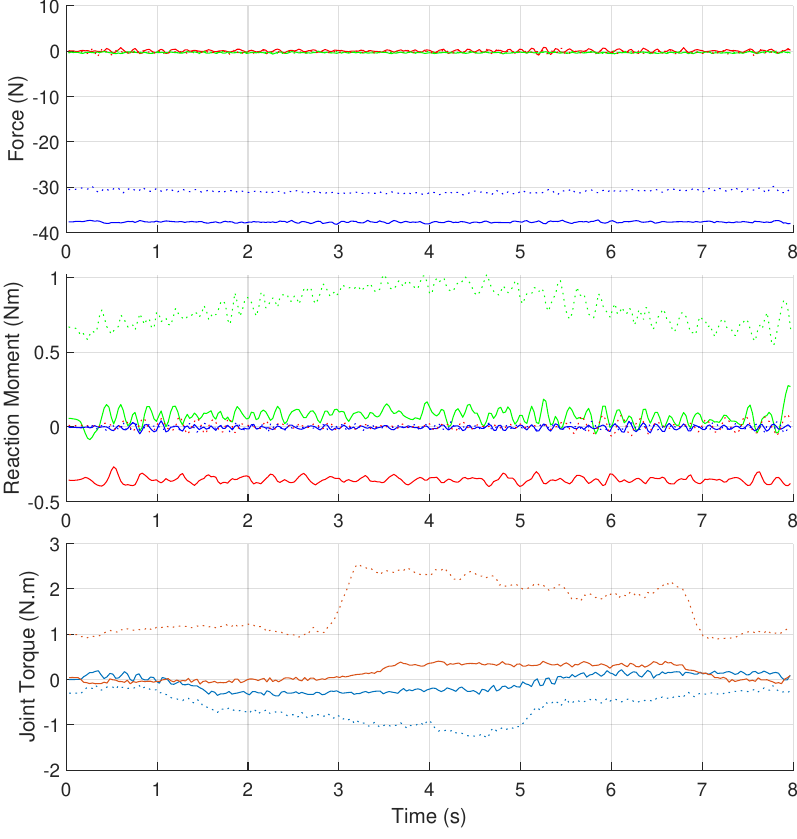}
    \caption{Forbal-5 Trajectory 2 results. The reaction forces and moments are shown as red (x), green (y) and blue (z), while joint torques are light blue (\(q_{11}\)) and brown (\(q_{22}\)). Solid line is balanced while dotted is unbalanced.}
    \label{fig:forbal5_traj2_results}
\end{figure}

The measured reaction forces/moments as well as the joint torques are shown in Figure \ref{fig:forbal5_traj2_results}. The joint torque for \(q_0\) was negligible as the motion is planar, and the pose error for \(_B p_\beta = 0\) and \(_O p_\gamma = 0\) is below \(10^{-3}\) rad at all times.

\subsubsection{Trajectory 3}
\begin{figure}[!ht]
    \centering
    \begin{subfigure}[b]{\linewidth}
    \includegraphics[trim=3.5cm 11.25cm 4cm 11cm,clip,width=\linewidth]{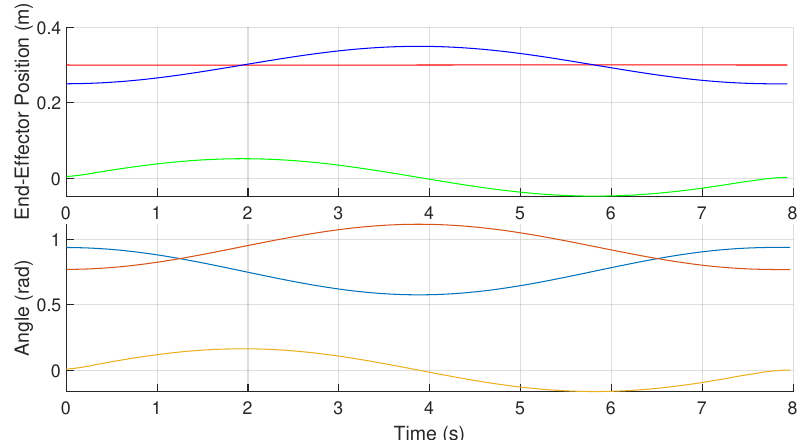}
    \caption{Forbal-5 Trajectory 3 motion profile.}
    \label{fig:forbal5_traj3}
    \end{subfigure}
    \begin{subfigure}[b]{\linewidth}
    \centering
        \includegraphics[trim=3.5cm 7.5cm 4cm 8cm,clip,width=\linewidth]{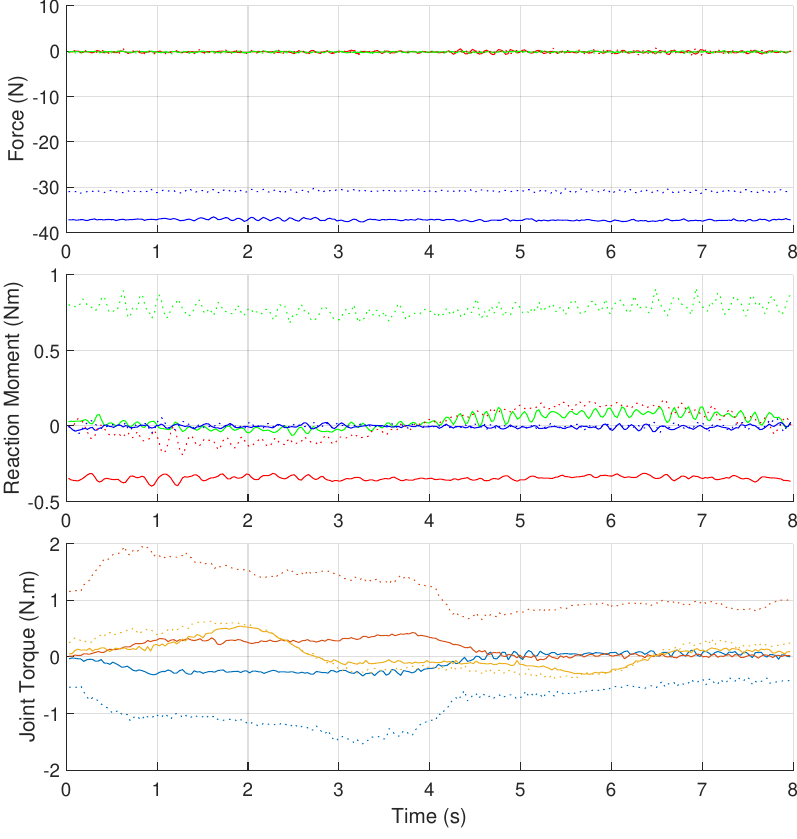}
        \caption{Trajectory 3 results averaged across 5 experiments.}
        \label{fig:forbal5_traj3_results}
    \end{subfigure}
    \caption{Forbal-5 Trajectory 3. Red (x), green (y) and blue (z) are positions and moments in the inertial frame, while light blue (\(q_{11}\)), brown (\(q_{22}\)) and yellow (\(q_0\)) are joint angles/torques.}
    \label{fig:forbal5_traj3_full}
\end{figure}

for Forbal-5 validated spatial motion under the inverse kinematic control, by tracing a circle in 8 seconds in the \(y_O z_O\) plane. The circle was centered on \(\icol{x_O & y_O & z_O} = \icol{0.3 & 0.0 & 0.3}\) with a radius of 0.1m, maintaining \(p_\beta = 0\) and \(p_\gamma = 0\) where the end-effector implement was aligned along the \(x_O\) axis and perpendicular to the \(x_O z_O\) plane. It was also sampled with 41 waypoints spaced at 0.2s increments. The motion profile and experimental results are shown in Figure \ref{fig:forbal5_traj3_full}.

\subsection{Discussion of Forbal-5 experimental results}
The major difference between Forbal-5 and Forbal-2 from a force balancing perspective is an increase in the end-effector payload. While the mechanical design of the EE motor with the additional 2 joints is designed to be as close to inline for which the counter mass placement is valid, in theory it is still not possible to perfectly force balance the design with the fixed counter mass positions of the base Forbal-2 design.

In practice, we observed that adding the counter masses still showed force balanced behavior without requiring gravity compensation from applying joint torques. Since the joints are not perfectly frictionless, static friction introduced from the Teflon bearings as well as the gears in the Dynamixel motors assisted in force balancing. As the required static friction torque contribution is quite low, this does not pose a significant issue with the dynamic motions of the manipulator while executing trajectories.

As the camera could only track planar trajectories, we also checked the position error through applying the forward kinematics on the reference and encoder-measured joint angles from the Dynamixel servomotors. However, similar to the Forbal-2 experiments, we found a consistent but negligible difference in position error metrics between the balanced and unbalanced in the order of 0.1 mm.

For the trajectories in the XZ plane, we observed greater improvements to the Forbal-2 results. Average Y reaction moments were reduced from 83-94\%, and the maximum Y reaction moment up to 86\%. Average absolute joint torques for \(q_{11}\) and \(q_{21}\) were reduced from 66-89\%, and the maximum absolute joint torque was reduced up to 84\%. For trajectory 3, we saw a reduction in \(q_0\) average absolute torque of 39\% as well. There was a greater difference between the balanced and unbalanced reaction moments and joint torques compared to Forbal-2, which could be attributed to the greater impact of counter mass balancing for a higher end effector mass. 

The planar reaction moments and joint torques were all reduced for the balanced manipulator, with the tradeoff of increased mass out of plane (y) reaction moment caused by increased mass for one side of the manipulator from asymmetric counter mass addition. These results demonstrate the practical application of the Forbal design for industrial applications, particularly where low speed and precision are required, and the improvements from force balancing are the most pronounced. 
\section{Conclusion}
\label{sec:conclusion}
In this research paper, we presented the novel design of a force balanced manipulator based on a planar five bar linkage. The geometric parameters and kinematic design were designed to maximize the workspace, and the dynamic parameters needed to force balance are determined by a mixture of link profile geometry and adjustable counter masses. 

Two variants were developed as a modular design: a planar 2-DOF variant called Forbal-2 and its extension to a full spatial 5-DOF design called Forbal-5. The first variant demonstrated the scientific validity through improvements in precision, reaction forces/moments, and joint torques from full balancing. The second variant extended the core scientific principle to a practical engineering application for industrial robotic manipulation.

The force balance conditions to fully force balance the manipulator were derived using the Linear Momentum with closed loop constraints, and different design considerations were justified through quantitative and qualitative analysis. We also derived the inverse kinematics of the manipulator by utilizing geometry and kinematics.

The manipulator variants were experimentally validated for 3-4 trajectories each, in both balanced and unbalanced configurations. We demonstrated that full balancing of the manipulator through calibrated counter masses improves the precision, reduces planar reaction moment and joint torques. This came with a tradeoff for a slight increase in lateral reaction moment due to asymmetry of the counter mass positioning. These improvements were more noticeable for slower trajectories where gravity compensation is a major component of reactions and joint torques, whereas for faster trajectories this improvement was less noticeable due to increased mass and inertia from the counter masses. For this reason, we recommend this design for precision industrial manipulation with a fixed payload where precision and fatigue are important considerations for the task.

Further validation of these designs through a motion capture system will allow more accurate quantification of the improvements to precision. The manipulator prototype was built from standard off-the-shelf mechanical components. The counter mass positions were also selected under inline assumptions for the links. Accurate modeling of non-inline center of mass offsets with adjustment of counter mass positions, as well as higher quality precision fabrication of mechanical components, can further improve the design through reducing modeling inaccuracies, unnecessary mass additions, and joint friction.

\section*{Acknowledgment}

The research leading to these results has received funding from the European Union's Horizon 2020 research and innovation programme under the Marie Sklodowska-Curie grant agreement no 101034319 and from the European Union — NextGenerationEU.

\ifCLASSOPTIONcaptionsoff
  \newpage
\fi



\bibliographystyle{IEEEtran}
\bibliography{IEEEabrv,references}
%



%

\vfill

\end{document}